\documentclass{article}

\usepackage[final]{neurips_2025}

\usepackage[utf8]{inputenc} 
\usepackage[T1]{fontenc}    
\usepackage{hyperref}       
\usepackage{url}            
\usepackage{booktabs}       
\usepackage{amsfonts}       
\usepackage{algorithm}
\usepackage{algpseudocode}
\usepackage{amsmath}
\usepackage{amssymb}
\usepackage{bm}
\usepackage{nicefrac}       
\usepackage{microtype}      
\usepackage{xcolor}         
\usepackage{multirow}
\usepackage{graphicx}
\usepackage{subcaption}
\usepackage{cleveref}
\usepackage{fancyvrb}
\usepackage{xcolor}
\usepackage{algorithm}
\usepackage{algpseudocode}
\usepackage{amsmath}

\title{FactsR: A Safer Method for Producing High Quality Healthcare Documentation}

\author{%
  Victor Petrén Bach Hansen \\
  Text Generation\\
  Corti.ai\\
  \texttt{vbh@corti.ai} \\
   \And
  Lasse Krogsb{\o}ll \\
  Clinical Product Specialist \\
  Corti.ai\\
  \texttt{lkr@corti.ai} \\
  \AND
  Jonas Lyngs{\o}\\
  Text Generation\\
  Corti.ai\\
  \texttt{lkr@corti.ai}\\
  \And
  Mathias Baltzersen \\
  Text Generation \\
  Corti.ai \\
  \texttt{mbn@corti.ai}
  \And
  Andreas Motzfeldt \\
  Text Generation \\
  Corti.ai \\
  \texttt{amo@corti.ai} \\
  \AND
  Kevin Pelgrims \\
  Text Generation \\
  Corti.ai \\
  kpe@corti.ai \\
  \And
  Lars Maal{\o}e \\
  Text Generation \\
  Corti.ai \\
  \texttt{lm@corti.ai} \\
}

\begin{document}

\maketitle

\begin{abstract}
There are now a multitude of AI-scribing solutions for healthcare promising the utilization of large language models for ambient documentation. However, these AI scribes still rely on one-shot, or few-shot prompts for generating notes after the consultation has ended, employing little to no reasoning. This risks long notes with an increase in hallucinations, misrepresentation of the intent of the clinician, and reliance on the proofreading of the clinician to catch errors. A dangerous combination for patient safety if vigilance is compromised by workload and fatigue. In this paper, we introduce a method for extracting salient clinical information in real-time alongside the healthcare consultation, denoted Facts, and use that information recursively to generate the final note. The FactsR method results in more accurate and concise notes by placing the clinician-in-the-loop of note generation, while opening up new use cases within real-time decision support.
\end{abstract}

\section{Introduction}
The clinical consultation is the cornerstone of a safe patient care journey, but it faces numerous threats. The healthcare professional must extract all relevant information from the patient within a limited time frame\citep{caldwell2019process} while simultaneously considering diagnostic and therapeutic options; a process that risks being imperfect. The resulting documentation must accurately and concisely encapsulate the consultation details, enabling seamless handovers between clinicians, and serve as legal documentation. The process of creating the note also serves as a time for reflection on the case, which makes it a sensitive phase for ensuring quality.

The manual generation of clinical notes is widely recognized as a time-intensive process\citep{baumann2018impact}. This documentation burden not only consumes valuable clinician time but may also compromise either the thoroughness of the record or the quality of patient interaction\citep{hart2025technology} in particular if threatened by time constraints or fatigue.

In response to this challenge, Ambient Scribes, systems leveraging automated abstractive summarization of medical dialogues, have seen rapid adoption, particularly following recent advances in large language models (LLMs)\citep{ma2024ambient}. These systems promise to reduce administrative overhead while maintaining or even enhancing the clarity and utility of clinical documentation\citep{tierney2024ambient,albrect2025enhancing}.

However, accurately summarizing healthcare consultations remains a significant challenge for LLMs\citep{vanveen2024adapted, kruse2025zeroshot}. Clinical conversations are characterized by complex medical terminology, diverse linguistic expressions, and a high degree of variability in dialogue structure. Crucially, clinically relevant information is often sparsely distributed and interwoven with extraneous or non-clinical dialogue, making it difficult to identify and extract salient content\citep{joshi2020drsummarize}.

This task is further complicated by the need for precise structuring of information according to clinical documentation standards (e.g., SOAP notes). Even human experts can struggle with this level of abstraction and organization. Evidence of this difficulty is reflected in performance on benchmark tasks such as the United States Medical Licensing Examination (USMLE), where annotators are required to identify key clinical concepts (rubric features) and correctly categorize them under the appropriate sections of a patient note\citep{yoon2016interrater,jin2021whatdisease}. These challenges underscore the complexity of training and evaluating LLMs for accurate and clinically coherent summarization.

Recent advances in LLM reasoning techniques offer promising avenues to address the limitations of both prompt engineering and finetuning in clinical summarization. Methods such as chain-of-thought prompting, retrieval-augmented generation, and modular reasoning pipelines enable models to decompose complex consultations into intermediate reasoning steps, improving accuracy and interpretability \cite{kojima2022large,wei2022chain}. By structuring the summarization process as a sequence of clinically-informed decisions, e.g., identifying relevant entities, mapping them to SOAP sections, and abstracting them into concise text, LLMs can achieve higher fidelity outputs while reducing reliance on brittle, large prompts, or domain-specific finetuning.

FactsR introduces a modular reasoning pipeline designed to produce accurate and concise clinical notes by orchestrating multiple LLMs. The design places the clinician at the center of this, achieving a degree of LLM-explainability by design. It operates in real time during healthcare consultations, continuously extracting relevant clinical information, performing structured clinical reasoning on the findings, and generating abstractive summaries tailored to the required documentation format. In conventional scribe systems, it is the clinician's job to review and edit the LLM-generated note, which is sometimes viewed as tedious and time consuming. FactsR is designed for ease of overview and understanding, by separating medical facts from formatting, phrasing, and allocation of information into a given note structure. In addition to greater clarity, it enables interactive engagement with intermediate clinical findings, allowing clinicians to refine and validate information on the fly, akin to glance at the notepad in classic note-taking. By extracting clinical findings in real time, FactsR unlocks vast opportunities for implementing AI-driven decision support systems at the point of care.

This paper makes three primary contributions: 
\begin{enumerate}
    \item it introduces FactsR, a novel framework that performs real-time clinical reasoning in tandem with the clinician during patient consultations;
    \item it proposes a model-based evaluation method tailored to assess the outputs of ambient scribe systems;
    \item and it presents a comparative analysis using the publicly available Primock57 benchmark, demonstrating the effectiveness of FactsR relative to a strong few-shot prompted ambient scribe baseline.
\end{enumerate}

Our findings indicate that the FactsR method significantly enhances the clinical relevance of generated notes by increasing the inclusion of pertinent information while simultaneously reducing extraneous content. This leads to notes that are both more accurate and more concise. Furthermore, when clinicians interact with the intermediate clinical findings produced by FactsR, we observe an additional improvement in both accuracy and conciseness, highlighting the value of clinician-guided refinement in the generation process.

\section{Motivation for FactsR}
Due to the unstructured nature of clinical dialogues and the computational expense associated with LLMs, most Ambient Scribe systems adopt a post-hoc generation approach, producing documentation only after the consultation has concluded. This strategy allows the model to process the entire conversation in a single pass, maximizing coherence by leveraging the full dialogue context during inference.

However, in addition to risking performance deterioration in long consultations, retrospective generation imposes challenges on structuring the input, as decomposing the raw transcript into intermediate reasoning steps post hoc is non-trivial\citep{mathur2023summqa,kirstein2025whats}. To compensate, these systems often rely on few-shot prompting to guide the model’s summarization behavior. While effective to some extent, this approach can be brittle and labor-intensive to scale, especially given the variability in clinical discourse. The example in Figure \ref{fig:template} illustrates a minimal prompt, comprising approximately 65 tokens, used to steer the generation process.

\begin{figure}
    \centering
{\small
\begin{verbatim}
Write a structured SOAP note based on the following clinical transcript.

S (Subjective): Patient’s reported symptoms, concerns, history.
O (Objective): Vitals, physical findings, labs, imaging.
A (Assessment): Diagnosis or differential.
P (Plan): Treatment, tests, follow-up.

Use clear, professional medical language like the following example:
[Insert example here]

Transcript:
[Insert transcript here]
\end{verbatim}
}
    \caption{Example of a simple prompt template that, given a transcript and examples, produces a SOAP note.}
    \label{fig:template}
\end{figure}

While minimal few-shot prompts offer a lightweight means of guiding LLM behavior, (in our experience) they typically yield suboptimal performance in the clinical summarization setting. Achieving high-quality outputs often necessitates elaborate few-shot prompts that exceed 1,000 tokens. As prompt complexity increases, making incremental adjustments while maintaining the global context of the complex healthcare encounter becomes increasingly fragile, small changes can propagate unintended effects across the output, reducing robustness, quality and maintainability.

One common strategy to overcome these limitations is finetuning the language model on domain-specific data\citep{anisuzzaman2024fine}. Although finetuned models generally outperform prompt-based approaches in constrained scenarios, they still exhibit diminished accuracy when confronted with complex, heterogeneous consultations. This suggests that neither few-shot prompting nor finetuning alone fully addresses the challenges inherent to the nuanced reasoning and contextual understanding required in clinical summarization tasks.

\section{FactsR}

\begin{algorithm}
\caption{Agentic Self-Refinement of Facts}
\begin{algorithmic}
\Require Transcript $W^{(j)}$, Existing Facts $F^{(\leq j)}$, Draft model $f_\theta$, Evaluator model $e_\phi$, Refinement model $r_\psi$, Max refinement steps $N_{\text{max}}$
\Ensure Refined set of facts $F^{(\leq j)}_{\text{new}}$

\State Initialize draft facts: $F^{(\leq j)}_{\text{draft}} \gets f_\theta(W^{(j)}, F^{(\leq j)})$

\ForAll{$f \in F^{(\leq j)}_{\text{draft}}$ \textbf{in parallel}}
    \State $f_{\text{current}} \gets f$
    \State $n \gets 0$
    \While{$n < N_{\text{max}}$}
        \State Evaluate $f_{\text{current}}$ with $e_\phi$: $e \gets e_{\phi}(f_{\text{current}}, W^{(j)})$
        \If{$f_{\text{current}}$ passes all criteria, $e$}
            \State Add $f_{\text{current}}$ to $F^{(\leq j)}_{\text{new}}$
            \State \textbf{break}
        \Else
            \State Refine: $f_{\text{current}} \gets r_\psi(f_{\text{current}}, e, W^{(j)})$
        \EndIf
        \State $n \gets n + 1$
    \EndWhile
\EndFor

\State \Return $F^{(\leq j)}_{\text{new}}$
\end{algorithmic}
\end{algorithm}

FactsR may be viewed as a rationale-based method for abstractive summarization. Rationales (or Facts) are extracted from a time sequence of the healthcare consultation. Each accumulated set of Facts are computed through a modular reasoning pipeline based on LLMs. After a healthcare consultation has ended, the final set of Facts are used to generate the final abstractive summarization.  

Let $V$ denote the vocabulary of tokens. Given a transcript $T = (t_1, t_2, \ldots, t_n)$ where each token $t_i \in V$, and $n = |T|$ is the total number of tokens in the transcript.

The goal is to extract a set of Facts $F = (f_1, f_2, \ldots, f_m)$, where $m$ is the total number of Facts present in the transcript. Fact extraction is performed by a set of large language models $f_\theta(\cdot)$, $e_{\phi}(\cdot)$, and $r_{\psi}(\cdot)$ parameterized by $\theta$, $\phi$, $\psi$\footnote{This is a technical report and we will not release the parameters of the LLMs. To test the FactsR method, you can get access on \hyperlink{https://www.corti.ai}{Corti.ai}.}. These large language models are used in combination with a self-refinement algorithm $\mathrm{Refine}(\cdot)$. FactsR is designed in such a way that $e_\phi$ can easily be extended to a set of individual criteria, each parameterized by smaller specialized models that excel at domain specific tasks. This algorithm avoids any adverse effects that may arise from streaming. 

Since the transcript represents a time-series (e.g., from a healthcare consultation), it is segmented into overlapping sliding windows of fixed length $w$. The complete set of such windows is defined as:

\begin{equation}
\mathcal{W} = \left\{ W^{(j)} = (t_j, t_{j+1}, \ldots, t_{j+w-1}) \,\middle|\, 1 \le j \le n - w + 1 \right\}.
\end{equation}

To compute the set of all Facts, we apply the model and refinement procedure iteratively or incrementally over the observed windowed context. Note that in practice, computing the Facts is not done on a per-token basis but rather on a more suitable update frequency window every $X$ received tokens in a way that balances compute budget with an acceptable level of periodical updates. The output of the algorithm after processing up to window $j$ is:

\begin{equation}
F^{(j)} = \mathrm{Refine} \left( f_\theta, e_{\phi}, r_{\psi}, F^{(\leq j)}, W^{(\leq j)}, N_{max} \right),
\end{equation}

where $W^{(\leq j)} = \left( W^{(1)}, W^{(2)}, \ldots, W^{(j)} \right)$ is the set of all windows up to index $j$. Finally, the complete set of extracted Facts is given by:
\begin{equation}
F = F^{(n - w + 1)}\ .    
\end{equation}

In order to produce the final clinical documentation, we use an instruction tuned language model $f_{\omega}$, parameterized by $\omega$, along with a shared clinical note template $T_{CN}$ (see simple example in Figure \ref{fig:template}), where $D_T = f_{\omega}(T,T_{CN})$ is a generated document from a standard Ambient Scribe and $D_F = f_{\omega}(F,T_{CN})$ the generated document from a FactsR Ambient Scribe. The template we use in our experiments is a general purpose template, designed to capture most medical information.

\section{Experiments}
\subsection{Dataset}
For the purpose of this analysis we focus on the the Primock57 dataset \citep{korfiatis2022primock57}\footnote{\url{https://github.com/babylonhealth/primock57}}, which is a publicly available benchmark dataset of 57 clinical encounters that includes manually transcribed audio of mock medical primary care consultations and corresponding physician-written clinical notes. We will use the physician-written notes as the gold standard for meaning, and use the meaning contained in the note for simulating a clinician-in-the-loop. This introduces circularity and limits conclusions, but also provides us with a useful benchmark.

\subsection{Evaluation}

The language in a healthcare consultation is complex making evaluating the output of ambient documentation with standard natural language processing metrics, like F1, BLEU, ROUGE, and METEOR, intractable\citep{navarro2025expert}. For example, if a patient consultation describes low blood sugar (hypoglycemia), two correct summarizations might be:
{\small
\begin{verbatim}
“The patient is exhibiting signs of hypoglycemia, which can result in dizziness, 
confusion, and in severe cases, loss of consciousness.”
\end{verbatim}
}
{\small
\begin{verbatim}
“Low blood sugar levels are present, which may lead to symptoms like confusion, 
fainting, or even coma if untreated.”
\end{verbatim}
}

whereas a wrong summary could be:
{\small
\begin{verbatim}
“The patient’s blood sugar is low, which helps prevent dizziness, confusion, 
and loss of consciousness.”
\end{verbatim}
}

Even though the last statement is wrong, metrics like BLEU and ROUGE might rank it higher because it shares many n-grams: \textit{"blood sugar is low"} and \textit{"dizziness, confusion, and loss of consciousness"}. Furthermore, METEOR and ROUGE might overvalue lexical overlap with the correct models, missing semantic reversal.

\subsubsection{Model-as-a-Judge}
Model-as-a-Judge is an emerging approach in machine learning for evaluating text generation systems, where a LLM is used to assess the quality of generated outputs in place of traditional automatic metrics or human annotators \citep{gao2024llmbasednlgevaluationcurrent,li2024llmsasjudgescomprehensivesurveyllmbased,yehudai2025surveyevaluationllmbasedagents}. Instead of relying on surface-level comparisons like BLEU or ROUGE, which measure lexical overlap, Model-as-a-Judge techniques prompt an LLM with the input, the generated response, and a reference, and ask it to rate or compare outputs based on dimensions like factual accuracy, coherence, helpfulness, or fidelity to the input intent. This paradigm aims to capture nuanced qualities of language, such as clinical correctness, logical consistency, or tone, that standard metrics often miss. When calibrated and prompted correctly, model judges have been shown to correlate well with human evaluations, offering a scalable and more semantically aware alternative for assessing generated text in domains like summarization, dialogue, and medical reasoning. However, Model-as-a-Judge has a tendency of being biased towards a set of predefined metrics not properly aligned with task at hand.

\subsubsection{Alignment based evaluation metrics}
In this evaluation, we use LLM alignment models as a Model-as-a-Judge. This leverages instruction-tuned or reinforcement-aligned language models to evaluate the quality of generated text. These alignment models are optimized to follow the semantic meanings in the Transcript, $T$, aligns with the output produced in the standard ambient scribe output, $D_T$, and the FactsR output, $D_F$. Finally, they evaluate how both $D_T$ and $D_F$ align with the gold note, $D_G$, written by a human clinician. Thus, this analysis does not analyze a translated definition of human preference but rather a direct semantic similarity in the source human annotated data sources, $T$ and $D_G$, with the machine-generated notes, $D_T$ and $D_F$. \citet{arora2025healthbenchevaluatinglargelanguage} presents an evaluation showing that Model-as-a-Judge overlaps with human expert annotations in the healthcare domain.

To simulate human involvement, we define $F_p$ and $F_{\hat{p}}$ as
\begin{align*}
    F_p &= \{ f \in F \mid f \not\subseteq D_G \} \quad \text{(facts from $F$ not aligning with $D_G$)} \\
    F_{\hat{p}} &= F_p \cup \{ s \in D_G \mid s \not\subseteq F \} \quad \text{($F_p$ plus parts of $D_G$ not covered by $F$)}
\end{align*}

In the real-world domain $F_{\hat{p}}$ can be processed, not from $D_G$, but rather from taking into account metadata, e.g., patient history, laboratory results, and scanning results, in order to further refine the documentation. FactsR can easily be extended to use this information as a part of its $\mathrm{Refine}(\cdot)$ algorithm.

The alignment model is a LLM turned into a multi-label classification model. Given a summary $x$, a set of segments $y_1, ..., y_n$, the model is tasked with:
\begin{enumerate}
    \item[] Among the segments $y_1, ..., y_n$, which are summarized by segment, $x$, if any?
\end{enumerate}
    
The model classifies the segments into two categories defined on a $[0, 1]$ scale as:
\begin{enumerate}
    \item[] 1: the summary conveys the same meaning as the source text.
    \item[] 0: the summary contradicts the source text or the summary contains information not present in the source text.
\end{enumerate}
Given two text-based sets (e.g. a segmented transcript and a segmented clinical note, or two different clinical notes), the alignment model then enables one to effectively construct a bipartite graph, semantically linking the individual items between them.

The alignment model therefore allows us to define three metrics tailored to the case of ambient documentation:
\begin{enumerate}
    \item \textbf{Completeness}: The proportion of $D_G$ represented in the generated note. A score of 100\% means all meaning in $D_G$ is captured; 0\% means none is represented. Since completeness is measured against $D_G$, and $D_G$ may contain content that is not grounded in the transcript, the maximum achievable completeness is limited by the groundedness (see below) of $D_G$. We therefore also present an \textbf{Adjusted Completeness} = Completeness / Groundedness (of $D_G$).
    \item \textbf{Conciseness}: The proportion of the generated note that is also found in $D_G$, reflecting clinical relevance. A score of 100\% means all generated content is relevant; 0\% means none is.
    \item \textbf{Groundedness}: The proportion of the statements in the note that represented in the transcript, $T$. This differs from hallucination rate because the reasons for discrepancies include human input in the gold-standard note as well as alignment model inaccuracies.
\end{enumerate}

These metrics thereby relate to how one would calculate recall and precision for the case of generated notes.

To compute these metrics, we use our alignment model to compare individual statements across the data sources. Each fact contains up to four discrete pieces of medical information (e.g., \textit{"diarrhea five times daily for 10 days"}), whereas $D_G$ is a condensed clinical note. To ensure accurate comparisons, we segment the notes into standalone sentences before applying the alignment model. 

The alignment model identifies which target statements are missing from the comparison document. It accepts partial alignments.

We tune our alignment model to be as strict (/unbiased) as possible by requiring that $D_G$ must align with itself (/$D_G$) with a 100\% \textbf{Completeness} and \textbf{Conciseness} score. If these scores are not 100\% it would indicate erroneous behavior of the alignment model and the segmentation of the notes.

\subsection{Results}
\begin{table}[!ht]
\centering
\begin{tabular}{lccccc}
\hline
\textbf{Approach} & \textbf{Completeness} &  \textbf{Conciseness} & \textbf{Groundedness} \\
\hline
$T \rightarrow D_T $ & 0.802 & 0.851 & 0.971 \\
$F \rightarrow D_F$ & 0.814 & 0.878 & 0.922 \\
$F_p \rightarrow D_p$ & 0.821 & 0.946 & 0.946 \\
$F_{\hat{p}} \rightarrow D_{\hat{p}}$ & 0.931 & 0.948 & 0.914 \\
$D_G$ & 1.0 & 1.0 & 0.922 \\
\hline
\end{tabular}
\caption{Evaluation of FactsR note generation process. We show Adjusted Completeness for all model-based results and Completeness for the gold-labeled notes.}
\label{tab:note_evaluation}
\end{table}

Table \ref{tab:note_evaluation} presents the quantitative evaluation of various approaches to note generation in the FactsR framework, comparing them on completeness, conciseness, and groundedness. The baseline method, Ambient Scribe ($T \rightarrow D_T$), shows the highest groundedness to the transcript (0.971), but this comes at the cost of completeness (0.802) and conciseness (0.851). This suggests that while the generated content remains close to the transcript, it diverges from the clinician’s intended note by failing to abstract or filter relevant information effectively. 

In contrast, the gold-labeled clinician notes ($D_G$) achieve perfect scores for completeness and conciseness (1.0), yet they exhibit the lowest groundedness (0.922). This highlights the clinician’s preference for generating a condensed and abstracted summary, rather than a literal reflection of the transcript, a key insight for modeling objectives. The perfect performance of completeness and conciseness is also an indication that the Model-as-a-Judge approach works well.

FactsR’s progression from $F \rightarrow D_F$ to $F_{\hat{p}} \rightarrow D_{\hat{p}}$ illustrates how iterative intervention strategies can shift generated notes closer to the clinician’s intent. Specifically, filtering out irrelevant facts ($F_p \rightarrow D_p$) yields improvements in both completeness (0.821) and conciseness (0.946), while a further step of simulating the addition of clinically meaningful, transcript-absent facts ($F_{\hat{p}} \rightarrow D_{\hat{p}}$) leads to a significant boost in completeness (0.931) and maintains high conciseness (0.948). Although this final step slightly reduces groundedness (0.914), since we introduce the clinicians own knowledge which might not be grounded in the transcript, it remains close to the clinician reference point, supporting the trade-off as clinically acceptable. The average number of removed Facts was 6.3 (equivalent to clicks in a user interface) and the average number added Facts was 4.1 per note (equivalent to dictated sentences in a user interface).

Post hoc clinician analysis of $D_{\hat{p}}$ revealed that much of the remaining incompleteness stemmed from overly strict alignment constraints imposed by the evaluation model, rather than true content omissions, favoring a conservative, less biased assessment over potentially misleading over-alignment.
Additionally, the choice of documentation template used, $T_{CN}$, also plays a significant role when generating the final clinical note as it ultimately decides what information is in enhanced or ignored in the source material. If there is a misalignment between the information we extract and the information we compare against, there is a natural ceiling to the performance we can achieve. We have purposefully chosen to not tailor a template specifically to the Primock57 consultation notes, but rather use one of Corti's general purpose templates, which attempt to cover most medical use-cases.

\section{Conclusion}
We introduce FactsR, a modular, real-time, clinician-in-the-loop method for generating high-quality clinical documentation through structured reasoning and iterative fact refinement. Our evaluation demonstrates that FactsR produces notes that are significantly more complete and concise than those generated by traditional few-shot ambient scribe approaches, while maintaining a high level of groundedness. Simulating clinician-guided interventions further improves note quality, bringing generated content closer to expert-authored documentation. These results highlight the value of decomposing the summarization task into interpretable reasoning steps and involving clinicians in the loop to improve both safety and clinical fidelity.

\newpage
\bibliography{references}

\begin{thebibliography}{22}
\providecommand{\natexlab}[1]{#1}
\providecommand{\url}[1]{\texttt{#1}}
\expandafter\ifx\csname urlstyle\endcsname\relax
  \providecommand{\doi}[1]{doi: #1}\else
  \providecommand{\doi}{doi: \begingroup \urlstyle{rm}\Url}\fi

\bibitem[Caldwell(2019)]{caldwell2019process}
Gordon Caldwell.
\newblock The process of clinical consultation is crucial to patient outcomes and safety: 10 quality indicators.
\newblock \emph{Clinical Medicine}, 19:\penalty0 503--506, 11 2019.

\bibitem[Baumann et~al.(2018)Baumann, Baker, and Elshaug]{baumann2018impact}
Lisa Baumann, Jannah Baker, and Adam Elshaug.
\newblock The impact of electronic health record systems on clinical documentation times: A systematic review.
\newblock \emph{Health Policy}, 122, 2018.

\bibitem[Hart et~al.(2025)Hart, Martin, Todd, and Hosking]{hart2025technology}
Graeme~K. Hart, Lorelle Martin, Julia Todd, and Nicole Hosking.
\newblock Technology-based challenges of informal clinical communication in an australian tertiary referral hospital: a survey-based assessment of user perspectives.
\newblock \emph{BMJ Open Quality}, 14, 2025.

\bibitem[Ma et~al.(2024)Ma, Liang, Shah, Smith, Jeong, Devon-Sand, Crowell, Delahaie, Hsia, Lin, Shanafelt, Pfeffer, Sharp, and Garcia]{ma2024ambient}
Stephen~P Ma, April~S Liang, Shreya~J Shah, Margaret Smith, Yejin Jeong, Anna Devon-Sand, Trevor Crowell, Clarissa Delahaie, Caroline Hsia, Steven Lin, Tait Shanafelt, Michael~A Pfeffer, Christopher Sharp, and Patricia Garcia.
\newblock Ambient artificial intelligence scribes: utilization and impact on documentation time.
\newblock \emph{Journal of the American Medical Informatics Association}, 32\penalty0 (2):\penalty0 381--385, 12 2024.

\bibitem[Tierney et~al.(2024)Tierney, Gayre, Hoberman, Mattern, Ballesca, Kipnis, Liu, and Lee]{tierney2024ambient}
Aaron Tierney, Gregg Gayre, Brian Hoberman, Britt Mattern, Mamuel Ballesca, Patricia Kipnis, Vincent Liu, and Kristine Lee.
\newblock Ambient artificial intelligence scribes to alleviate the burden of clinical documentation.
\newblock \emph{NEJM Catalyst}, 5, 02 2024.

\bibitem[Albrecht et~al.(2025)Albrecht, Shanks, Shah, Hudson, Thompson, Filardi, Wright, Ator, and Smith]{albrect2025enhancing}
Michael Albrecht, Denton Shanks, Tina Shah, Taina Hudson, Jeffrey Thompson, Tanya Filardi, Kelli Wright, Gregory~A Ator, and Timothy~Ryan Smith.
\newblock Enhancing clinical documentation with ambient artificial intelligence: a quality improvement survey assessing clinician perspectives on work burden, burnout, and job satisfaction.
\newblock \emph{JAMIA Open}, 8\penalty0 (1), 02 2025.

\bibitem[Van~Veen et~al.(2024)Van~Veen, Van~Uden, Blankemeier, Delbrouck, Aali, Bluethgen, Pareek, Polacin, Reis, Seehofnerová, Rohatgi, Hosamani, Collins, Ahuja, Langlotz, Hom, Gatidis, Pauly, and Chaudhari]{vanveen2024adapted}
D.~Van~Veen, C.~Van~Uden, L.~Blankemeier, J.~B. Delbrouck, A.~Aali, C.~Bluethgen, A.~Pareek, M.~Polacin, E.~P. Reis, A.~Seehofnerová, N.~Rohatgi, P.~Hosamani, W.~Collins, N.~Ahuja, C.~P. Langlotz, J.~Hom, S.~Gatidis, J.~Pauly, and A.~S. Chaudhari.
\newblock Adapted large language models can outperform medical experts in clinical text summarization.
\newblock \emph{Nature Medicine}, 30\penalty0 (4):\penalty0 1134--1142, 2024.

\bibitem[Kruse et~al.(2025)Kruse, Hu, Derby, Wu, Stonbraker, Yao, Wang, Goldberg, and Gao]{kruse2025zeroshot}
Maya Kruse, Shiyue Hu, Nicholas Derby, Yifu Wu, Samantha Stonbraker, Bingsheng Yao, Dakuo Wang, Elizabeth Goldberg, and Yanjun Gao.
\newblock Zero-shot large language models for long clinical text summarization with temporal reasoning, 2025.
\newblock URL \url{https://arxiv.org/abs/2501.18724}.

\bibitem[Joshi et~al.(2020)Joshi, Katariya, Amatriain, and Kannan]{joshi2020drsummarize}
Anirudh Joshi, Namit Katariya, Xavier Amatriain, and Anitha Kannan.
\newblock Dr. summarize: Global summarization of medical dialogue by exploiting local structures.
\newblock In \emph{Findings of the Association for Computational Linguistics: EMNLP 2020}, pages 3755--3763. Association for Computational Linguistics, November 2020.

\bibitem[Park et~al.(2016)Park, Hyderi, Bordage, Xing, and Yudkowsky]{yoon2016interrater}
Yoon~Soo Park, Abbas Hyderi, Georges Bordage, Kuan Xing, and Rachel Yudkowsky.
\newblock Inter-rater reliability and generalizability of patient note scores using a scoring rubric based on the usmle step-2 cs format.
\newblock \emph{Advances in Health Sciences Education}, 21:\penalty0 1--13, 10 2016.

\bibitem[Jin et~al.(2021)Jin, Pan, Oufattole, Weng, Fang, and Szolovits]{jin2021whatdisease}
Di~Jin, Eileen Pan, Nassim Oufattole, Wei-Hung Weng, Hanyi Fang, and Peter Szolovits.
\newblock What disease does this patient have? a large-scale open domain question answering dataset from medical exams.
\newblock \emph{Applied Sciences}, 11:\penalty0 6421, 07 2021.

\bibitem[Kojima et~al.(2022)Kojima, Gu, Reid, Matsuo, and Iwasawa]{kojima2022large}
Takeshi Kojima, Shixiang~Shane Gu, Machel Reid, Yutaka Matsuo, and Yusuke Iwasawa.
\newblock Large language models are zero-shot reasoners.
\newblock In \emph{Proceedings of the 36th International Conference on Neural Information Processing Systems}, NIPS 22, 2022.

\bibitem[Wei et~al.(2022)Wei, Wang, Schuurmans, Bosma, Ichter, Xia, Chi, Le, and Zhou]{wei2022chain}
Jason Wei, Xuezhi Wang, Dale Schuurmans, Maarten Bosma, Brian Ichter, Fei Xia, Ed~H. Chi, Quoc~V. Le, and Denny Zhou.
\newblock Chain-of-thought prompting elicits reasoning in large language models.
\newblock In \emph{Proceedings of the 36th International Conference on Neural Information Processing Systems}, NIPS 22, 2022.

\bibitem[Mathur et~al.(2023)Mathur, Rangreji, Kapoor, Palavalli, Bertsch, and Gormley]{mathur2023summqa}
Yash Mathur, Sanketh Rangreji, Raghav Kapoor, Medha Palavalli, Amanda Bertsch, and Matthew Gormley.
\newblock {S}umm{QA} at {MEDIQA}-chat 2023: In-context learning with {GPT}-4 for medical summarization.
\newblock In \emph{Proceedings of the 5th Clinical Natural Language Processing Workshop}, pages 490--502. Association for Computational Linguistics, July 2023.

\bibitem[Kirstein et~al.(2025)Kirstein, Lima~Ruas, and Gipp]{kirstein2025whats}
Frederic~Thomas Kirstein, Terry Lima~Ruas, and Bela Gipp.
\newblock What`s wrong? refining meeting summaries with {LLM} feedback.
\newblock In Owen Rambow, Leo Wanner, Marianna Apidianaki, Hend Al-Khalifa, Barbara~Di Eugenio, and Steven Schockaert, editors, \emph{Proceedings of the 31st International Conference on Computational Linguistics}, pages 2100--2120, Abu Dhabi, UAE, January 2025. Association for Computational Linguistics.

\bibitem[Anisuzzaman et~al.(2024)Anisuzzaman, Malins, Friedman, and Attia]{anisuzzaman2024fine}
DM~Anisuzzaman, JG~Malins, PA~Friedman, and ZI~Attia.
\newblock Fine-tuning large language models for specialized use cases.
\newblock \emph{Mayo Clinic Proceedings: Digital Health}, 3\penalty0 (1):\penalty0 100184, Nov 29 2024.

\bibitem[Papadopoulos~Korfiatis et~al.(2022)Papadopoulos~Korfiatis, Moramarco, Sarac, and Savkov]{korfiatis2022primock57}
Alex Papadopoulos~Korfiatis, Francesco Moramarco, Radmila Sarac, and Aleksandar Savkov.
\newblock (in press): Primock57: A dataset of primary care mock consultations.
\newblock In \emph{Proceedings of the 60th Annual Meeting of the Association for Computational Linguistics}, 2022.

\bibitem[Fraile~Navarro et~al.(2025)Fraile~Navarro, Coiera, Hambly, Triplett, Asif, Susanto, Chowdhury, Lorenzo, Dras, and Berkovsky]{navarro2025expert}
David Fraile~Navarro, Enrico Coiera, Thomas Hambly, Zoe Triplett, Nahyan Asif, Anindya Susanto, Anamika Chowdhury, Amaya Lorenzo, Mark Dras, and Shlomo Berkovsky.
\newblock Expert evaluation of large language models for clinical dialogue summarization.
\newblock \emph{Scientific Reports}, 15, 01 2025.

\bibitem[Gao et~al.(2024)Gao, Hu, Ruan, Pu, and Wan]{gao2024llmbasednlgevaluationcurrent}
Mingqi Gao, Xinyu Hu, Jie Ruan, Xiao Pu, and Xiaojun Wan.
\newblock Llm-based nlg evaluation: Current status and challenges, 2024.
\newblock URL \url{https://arxiv.org/abs/2402.01383}.

\bibitem[Li et~al.(2024)Li, Dong, Chen, Su, Zhou, Ai, Ye, and Liu]{li2024llmsasjudgescomprehensivesurveyllmbased}
Haitao Li, Qian Dong, Junjie Chen, Huixue Su, Yujia Zhou, Qingyao Ai, Ziyi Ye, and Yiqun Liu.
\newblock Llms-as-judges: A comprehensive survey on llm-based evaluation methods, 2024.
\newblock URL \url{https://arxiv.org/abs/2412.05579}.

\bibitem[Yehudai et~al.(2025)Yehudai, Eden, Li, Uziel, Zhao, Bar-Haim, Cohan, and Shmueli-Scheuer]{yehudai2025surveyevaluationllmbasedagents}
Asaf Yehudai, Lilach Eden, Alan Li, Guy Uziel, Yilun Zhao, Roy Bar-Haim, Arman Cohan, and Michal Shmueli-Scheuer.
\newblock Survey on evaluation of llm-based agents, 2025.
\newblock URL \url{https://arxiv.org/abs/2503.16416}.

\bibitem[Arora et~al.(2025)Arora, Wei, Hicks, Bowman, Quiñonero-Candela, Tsimpourlas, Sharman, Shah, Vallone, Beutel, Heidecke, and Singhal]{arora2025healthbenchevaluatinglargelanguage}
Rahul~K. Arora, Jason Wei, Rebecca~Soskin Hicks, Preston Bowman, Joaquin Quiñonero-Candela, Foivos Tsimpourlas, Michael Sharman, Meghan Shah, Andrea Vallone, Alex Beutel, Johannes Heidecke, and Karan Singhal.
\newblock Healthbench: Evaluating large language models towards improved human health, 2025.
\newblock URL \url{https://arxiv.org/abs/2505.08775}.

\end{thebibliography}
\end{document}